\documentclass[review, a4paper]{cas-sc}

\usepackage[numbers,longnamesfirst]{natbib}

\usepackage{amsfonts}       
\usepackage{nicefrac}       
\usepackage{microtype}      
\usepackage{graphicx}
\usepackage{amsmath}
\usepackage[rightComments=false]{algpseudocodex}
\usepackage{algorithm}
\usepackage{multirow}
\usepackage{soul}
\usepackage[dvipsnames]{xcolor}
\usepackage{color}
\usepackage{flafter} 
\usepackage{tabularx}
\usepackage{float}

\colorlet{ChangesColor}{SkyBlue!0}

\sethlcolor{ChangesColor}

\makeatletter
\renewcommand\appendix{\par
  \setcounter{section}{0}%
  \setcounter{subsection}{0}%
  \setcounter{equation}{0}%
  \setcounter{table}{0}
  \setcounter{figure}{0}
  \gdef\theequation{\@Alph\c@section.\arabic{equation}}%
  \gdef\thefigure{\@Alph\c@section.\arabic{figure}}%
  \gdef\thetable{\@Alph\c@section.\arabic{table}}%
  \gdef\thesection{\appendixname \ \@Alph\c@section}%
  \@addtoreset{equation}{section}%
  \@addtoreset{table}{section}
  \@addtoreset{figure}{section}
}
\makeatother

\begin{document}
\let\WriteBookmarks\relax
\def\floatpagepagefraction{1}
\def\textpagefraction{.001}

\shorttitle{Positive-Unlabelled Learning for Dietary Restriction Genetics}

\shortauthors{Jorge Paz-Ruza et al.}

\title [mode = title]{Positive-Unlabelled Learning for Identifying New Candidate Dietary Restriction-related Genes among Ageing-related Genes}                      



%
\author[1]{Jorge Paz-Ruza}[orcid=0000-0002-7869-1070]
\cormark[1]
\ead[url]{j.ruza@udc.es}
\credit{Conceptualization, Methodology, Software, Investigation, Writing - Original Draft, Writing – Review \& Edit}

\author[2]{Alex A. Freitas}[orcid=0000-0001-9825-4700]
\ead[url]{A.A.Freitas@kent.ac.uk}
\credit{Conceptualization, Methodology, Writing - Original Draft, Writing - Review \& Editing, Supervision}

\author[1]{Amparo Alonso-Betanzos}[orcid=0000-0003-0950-0012]
\ead[url]{amparo.alonso.betanzos@udc.es}
\credit{Conceptualization, Writing - Review \& Editing, Supervision}

\author[1]{Bertha Guijarro-Berdiñas}[orcid=0000-0001-8901-5441]
\ead[url]{berta.guijarro@udc.es}
\credit{Conceptualization, Writing - Review \& Editing, Supervision}

\affiliation[1]{organization={LIDIA Group, CITIC, Universidade da Coruña},
            addressline={Campus de Elviña s/n}, 
            city={A Coruña},
            citysep={}, 
            postcode={15071}, 
            country={Spain}}

\affiliation[2]{organization={School of Computing, University of Kent},
            city={Canterbury},
            citysep={}, 
            postcode={CT2 7FS},
            country={United Kingdom}}

\cortext[1]{Corresponding author}



\begin{abstract}
Dietary Restriction (DR) is one of the most popular anti-ageing interventions; recently, Machine Learning (ML) has been explored to identify potential DR-related genes among ageing-related genes, aiming to minimize costly wet lab experiments needed to expand our knowledge on DR. \hl{However, to train a model from positive (DR-related) and negative (non-DR-related) examples, the existing ML approach naively labels genes without known DR relation as negative examples, assuming that lack of DR-related annotation for a gene represents evidence of absence of DR-relatedness, rather than absence of evidence. This hinders the reliability of the negative examples (non-DR-related genes) and the method's ability to identify novel DR-related genes.} This work introduces a novel gene prioritisation method based on the two-step Positive-Unlabelled (PU) Learning paradigm: using a similarity-based, KNN-inspired approach, our method first selects reliable negative examples among the genes without known DR associations. Then, these reliable negatives and all known positives are used to train a classifier that effectively differentiates DR-related and non-DR-related genes, which is finally employed to generate a more reliable ranking of promising genes for novel DR-relatedness. \hl{Our method significantly outperforms ($p<0.05$) the existing state-of-the-art approach in three predictive accuracy metrics with up to $\sim40\%$ lower computational cost in the best case, and we identify 4 new promising DR-related genes (PRKAB1, PRKAB2, IRS2, PRKAG1), all with evidence from the existing literature supporting their potential DR-related role.}
\end{abstract}



\begin{keywords}
 Machine Learning \sep Positive-Unlabelled Learning \sep Bioinformatics \sep Ageing \sep Dietary Restriction 
\end{keywords}

\maketitle

\section{Introduction}

Ageing is a biological process characterized by a progressive decline in physiological function and increased susceptibility to age-related diseases. As it is a complex phenomenon influenced by both genetic and environmental factors \cite{Lopez-Otin2016}, understanding the genetic basis of ageing is crucial for deciphering its underlying mechanisms and developing methods to promote healthy ageing. The scientific community has invested a great amount of research efforts into understanding the biological processes involved in ageing; particularly, genetic studies have identified numerous genes and pathways associated with ageing, offering insights into potential targets for anti-ageing therapeutic interventions \cite{partridge2010new, guarente2000genetic, melzer2020genetics}.

One of the most promising and studied approaches to extend lifespan and delay the onset of age-related diseases is Dietary Restriction (DR), which involves reducing nutrient intake (and typically, calorie intake) without causing malnutrition \cite{most2017calorie} and has been shown to extend lifespan and improve long-term health in various model organisms \cite{das2004caloric}. By modulating various cellular pathways, such as insulin signalling, sirtuin activation, or autophagy induction \cite{kirk2009dietary, wood2004sirtuin, bergamini2007role}, DR promotes cellular stress resistance and metabolic efficiency, reducing the risk of age-related pathologies such as cardiovascular disease, cancer, and neurodegeneration \cite{lopez2016calorie, de2022calorie}.

The research efforts in ageing and other biomedical areas greatly increased the magnitude and complexity of available biological data; as a solution, Machine Learning (ML) has emerged as a powerful tool to facilitate the analysis of large-scale biological datasets and uncover hidden patterns \cite{angermueller2016deep, shastry2020machine}. ML techniques have been widely applied in ageing-related research, including the prediction of lifespan of model organisms, the identification of molecular signatures of ageing, and the association of metabolic pathways with ageing-related diseases  \cite{kern2024machine, fabris2017review}. 

In the particular topic of DR, Magdaleno et al. \cite{vega2022machine} recently employed ML to classify ageing-related genes into DR-related and non-DR-related genes, in order to identify candidate DR-related genes among ageing-related genes not currently annotated as DR-related. \hl{Utilizing various biological features, such as pathway information from PathDIP \mbox{\cite{rahmati2017pathdip}}, Gene Ontology (GO) terms \mbox{\cite{ashburner2000gene}}, KEGG pathways \mbox{\cite{kanehisa2000kegg}}, or coexpression data, they trained decision tree-based ensemble classifiers under a binary classification task to ultimately produce a ranking of promising DR-related genes for wet-lab verification. 

However, the pipeline described in \mbox{\cite{vega2022machine}} holds a significant limitation: to provide training examples with binary labels to the classifier, Magdaleno et al. assumed that all ageing-related genes without experimental evidence of DR-relatedness can be considered as non-DR-related, i.e. labelled as negative training examples. As the authors acknowledge, an absence of evidence does not equate to evidence of absence of DR-relatedness, meaning the classifier was trained with an unknown amount of incorrectly labelled negative examples, hindering its learning and therefore its ability to correctly identify new DR-related genes.}

This type of data, commonly referred to as Positive-Unlabelled (PU) data, corresponds to cases where a subset of the examples are labelled as known positives and, while the rest is unlabelled and comprises both negative and positive examples \cite{elkan2008learning}. This is common in bioinformatics due to the cost of obtaining annotations through wet-lab experimentation \cite{li2022positive}, but ignoring unlabelled data or treating it as negative (as usual in the literature) leads to biased and suboptimal models in many ML tasks \cite{kiryo2017positive}. As a solution, PU Learning is an ML paradigm specifically designed for improving the quality and predictive power of classifiers in settings that involve PU data, acknowledging unlabelled examples as such throughout the ML pipeline \cite{bekker2020learning}.

In this work, we propose a PU Learning method to enhance the prediction of novel candidate DR-related genes among ageing-related genes. \hl{Our approach addresses the limitations of the Magdaleno et al.'s \mbox{\cite{vega2022machine}} existing methodology by properly accounting for the unlabelled data and exploiting its potential to improve the quality of predictions. Specifically, we propose a similarity-based two-step PU Learning strategy to improve the training process and the predictive power of classifiers. The proposed method has surpassed the performance of the state-of-the-art non-PU Learning method in our target task of predicting novel DR-related genes among ageing-related genes.} 
In addition, we use this PU learning methodology to generate a more reliable list of top candidate genes for novel DR-relatedness supported by evidence in the literature of the field.

We underline three main contributions of this work:

\begin{itemize}
    \item \hl{Firstly, the design, implementation and experimental validation of a PU learning method for gene prioritisation, and its application to improve the identification of DR-related genes among ageing-related genes. In this task, our method significantly outperformed the existing state-of-the-art (non-PU) method in all three used predictive performance metrics, on real-world ageing-related gene datasets, while reducing the required computational overhead to train and employ the model for DR-related gene identification.}
    \item \hl{Secondly, the novel use of the proposed PU Learning method to produce a more reliable list of top candidate DR-related genes, compared to Magdaleno et al.'s \mbox{\cite{vega2022machine}} state-of-the-art approach, owing to our method's superior performance in computational experiments.}
    \item Lastly, a curation of the relevant literature, which identified evidence supporting the potential DR-relatedness of our method's top candidate genes, further motivating wet-lab experiments that could validate the predicted DR-relatedness of the proposed genes.
\end{itemize}

The remainder of this paper is structured as follows. \hl{In Section \mbox{\ref{sec:background}} we discuss our target task and existing methodology for identifying new DR-related genes, highlight its limitations, and introduce basic notions of the PU Learning ML paradigm and the need for it in the task.} Section \ref{sec:proposal} presents our proposed PU Learning method, detailing how it enhances the training process to improve classifiers' power. In Section \ref{sec:setup}, we describe the experimental setup, including features, classifiers and evaluation metrics. Section \ref{sec:results} presents the results of our experiments, comparing our PU Learning approach against the state-of-the-art non-PU approach. Section \ref{sec:conclusions} concludes the paper and discusses avenues for future research.
\section{Background}
\label{sec:background}

\hl{In this section we formalize the task of interest, cover the existing ML-based approach used to solve it and its limitations, and introduce basic concepts of the PU Learning paradigm in ML and its need for training classifiers in this task.}

\subsection{Task Formulation}
\label{sec:task}
The task at hand is to find new candidate genes related to DR among ageing-related genes. Formally, let $\mathcal{G}_{\text{AGE}}$ be the set of known ageing-related genes, each associated with a vector of $n$ biological features $\mathbf{x}_g = (f_1, f_2, ..., f_n)$.
We assume there exists a subset of genes $\mathcal{G}_{\text{DR}\cap\text{AGE}^+}$ involved in DR used as anti-ageing intervention. 
Among these, a smaller subset of genes $\mathcal{G}_{\text{DR}\cap\text{AGE}}$ have experimental evidence of DR-relatedness. Reasonably, not all ageing-related genes have a relation to DR, and not all DR-related genes have been identified experimentally, and therefore: 

\begin{equation}
    \mathcal{G}_{\text{DR}\cap\text{AGE}} \subset \mathcal{G}_{\text{DR}\cap\text{AGE}^+} \subset \mathcal{G}_{\text{AGE}}
\end{equation}

The objective is to identify ageing-related genes $g^*$ without known evidence of DR-relatedness but with high probability of actually being DR-related, such that:

\begin{equation}
g^* = \text{argmax}_{g \in \mathcal{G}_{\text{AGE}} \setminus \mathcal{G}_{\text{DR}\cap\text{AGE}}} Pr(g \in \mathcal{G}_{\text{DR}\cap\text{AGE}^+})
\label{eq:topcandidates}
\end{equation}

\noindent where $g^*$ are the genes most likely to be found to be DR-related in future wet-lab experiments, constituting a task of knowledge discovery, i.e. finding the subset $\mathcal{G}_{\text{DR}\cap\text{AGE}^+} \setminus \mathcal{G}_{\text{DR}\cap\text{AGE}}$.

A general solution of this task is to find a model \( \Phi: \mathcal{G}_{\text{AGE}} \rightarrow [0, 1] \) that assigns a score $\Phi(g)$ to each gene $g$ as a measure of its DR-relatedness. Since the only known information is which genes already belong to $\mathcal{G}_{\text{DR}\cap\text{AGE}}$, where trivially $Pr(g \in \mathcal{G}_{\text{DR}\cap\text{AGE}^+}) = 1$, the task is surrogated to a binary classification problem: the positive class are those genes that belong to \( \mathcal{G}_{\text{DR}\cap\text{AGE}^+} \), and the negative class consists of the remaining ageing-related genes (\( \mathcal{G}_{\text{AGE}} \setminus \mathcal{G}_{\text{DR}\cap\text{AGE}^+} \)).

An important detail of this formulation is that the real positive class is not composed solely of genes with known association with DR, but also with undiscovered DR-relatedness, i.e. that belong to the unknown set $\mathcal{G}_{\text{DR}\cap\text{AGE}^+} \setminus \mathcal{G}_{\text{DR}\cap\text{AGE}}$. Similarly, the actual negative class is not the set of genes without known DR-relatedness ($\mathcal{G}_{\text{AGE}} \setminus \mathcal{G}_{\text{DR}\cap\text{AGE}}$), but the set of genes without real DR-relatedness ($\mathcal{G}_{\text{AGE}} \setminus \mathcal{G}_{\text{DR}\cap\text{AGE}^+}$).

The model $\Phi$ to be designed is then a binary classifier that, given a gene's features, outputs the probability of it belonging to the positive class, such that:

\begin{equation}
    \Phi(g) = \begin{cases}
        x \in [0.5, 1] & g \in \mathcal{G}_{\text{DR}\cap\text{AGE}^+} \\
        x \in [0, 0.5) & g \in \mathcal{G}_{\text{AGE}} \setminus \mathcal{G}_{\text{DR}\cap\text{AGE}^+}
        
    \end{cases}
\end{equation}

Since $\Phi(g)$ can also be interpreted as a probability of DR-relatedness, it approximates the original goal, as seen in Equation \ref{eq:aprox}: the genes without known DR-relatedness given highest scores by $\Phi$ should be the most promising candidates for DR-relatedness.

\begin{equation}
\text{argmax}_{g \in \mathcal{G}_{\text{AGE}} \setminus \mathcal{G}_{\text{DR}\cap\text{AGE}}} \Phi(g)  \approx \text{argmax}_{g \in \mathcal{G}_{\text{AGE}} \setminus \mathcal{G}_{\text{DR}\cap\text{AGE}}} Pr(g \in \mathcal{G}_{\text{DR}\cap\text{AGE}^+}) \approx \mathcal{G}_{\text{DR}\cap\text{AGE}^+} \setminus \mathcal{G}_{\text{DR}\cap\text{AGE}}
\label{eq:aprox}
\end{equation}

\begin{figure}[h]
    \centering
    \includegraphics[width=\textwidth]{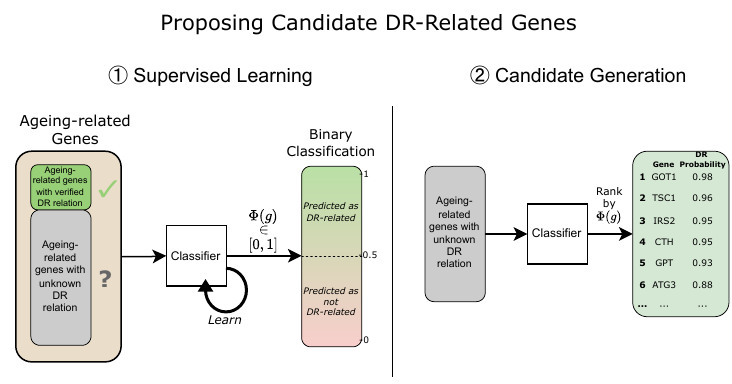}
    \caption{General overview of the two-step modelling to solve the task for proposing potential novel DR-related genes among ageing-related genes.}
    \label{fig:task}
\end{figure}

Figure \ref{fig:task} summarizes this discovery task as two steps: 1) surrogating the task to a binary classification to achieve a model which identifies DR-related genes among a set of ageing-related genes, and 2) using the model's predictions on ageing-related genes without known DR-relatedness to generate a ranking of promising, undiscovered DR-related genes. 

\subsection{\hl{Existing Methodology for DR-Related Gene Identification}}
\label{sec:existing}

\hl{Recently, Magdaleno et al. \mbox{\cite{vega2022machine}} explored the task of predicting novel DR-relatedness among ageing-related genes using ML techniques, showing promising results. 
By the time of our research, Magdaleno et al.'s work is the only work for identifying novel DR-related genes among ageing-related genes. In addition they used state-of-the-art classifiers for tabular data (decision tree ensembles)  \mbox{\cite{grinsztajn2022tree}}. Hence, their approach is the state-of-the-art for this task.}

To characterize each gene, they considered a variety of biological features: PathDIP gene-pathway interactions \cite{rahmati2017pathdip}, KEGG pertinence \cite{kanehisa2000kegg} and influence \cite{fabris2016new} descriptors, protein-protein interaction (PPI) \cite{oughtred2019biogrid} adjacency and graph metrics, hierarchical Gene Ontology (GO) terms \cite{ashburner2000gene}, expression information in tissues from GTEx \cite{gtex2020gtex}, gene co-expression data \cite{van2015genefriends}, and protein descriptors \cite{rainer2017ensdb}. These features were analyzed individually and collectively as potential predictors for DR-relatedness of ageing-related genes. 

In terms of classification algorithms, Magdaleno et al. used standard decision tree-based ensemble algorithms for their state-of-the-art performance in tabular data \cite{grinsztajn2022tree}, experimenting with Balanced Random Forest (BRF) \cite{chen04using}, XGBoost \cite{chen2015xgboost}, Easy Ensemble Classifier \cite{liu2008exploratory}, and CatBoost \cite{prokhorenkova2018catboost}, and employing a nested cross-validation to evaluate all combinations of features and classifiers. Interestingly, their results indicated that combining different types of biological features did not necessarily improve performance.

Two best-performing combinations of a feature type and a classifier were identified (\{PathDIP, CatBoost\} and \{GO, BRF\}), and then used to predict the DR-relatedness of genes without known DR association and propose a ranking of the most promising candidates.

A fundamental element of the methodology used by Magdaleno et al. \cite{vega2022machine} lies in the assumption that all genes without known DR relation can be considered negative examples during training. This is, the model is taught to predict as

\begin{equation}
    \Phi(g) = 0 \;\;\; \forall g \in \mathcal{G}_{\text{AGE}} \setminus \mathcal{G}_{\text{DR}\cap\text{AGE}} 
\end{equation}

\noindent but, by the formulation from Section \ref{sec:task}, this is equivalent to teaching the model that

\begin{equation}
\begin{gathered}
    Pr(g \in \mathcal{G}_{\text{DR}\cap\text{AGE}^+})=0 \;\;\; \forall g \in \mathcal{G}_{\text{AGE}} \setminus \mathcal{G}_{\text{DR}\cap\text{AGE}}  \\ \mathcal{G}_{\text{DR}\cap\text{AGE}^+} \setminus \mathcal{G}_{\text{DR}\cap\text{AGE}} = \emptyset
\end{gathered} 
\end{equation}

\noindent \hl{which defeats the purpose of the discovery task: retrieving with the ML algorithm the undiscovered DR-related genes that exist in the set of genes without known DR-relatedness, i.e. finding $\mathcal{G}_{\text{DR}\cap\text{AGE}^+} \setminus \mathcal{G}_{\text{DR}\cap\text{AGE}}$. Ultimately, this can lead to a lower performance of the ML model and, in consequence, a less reliable set of candidate genes for DR-relatedness. 

In this work, we show the design and usage of a more sophisticated labelling algorithm of training examples based on PU Learning, making it aware that unlabelled examples are not necessarily negative (i.e. that some genes without known DR-relatedness may actually be DR-related), can overcome the aforementioned limitation of Magdaleno et al.'s state-of-the-art approach and improve the identification of novel DR-related genes without adding computational overhead (by comparison with Magdaleno et al.'s approach).}

\subsection{\hl{Essential Notions of PU Learning}}

In the context of semi-supervised learning, PU Learning is an ML paradigm designed for scenarios where, rather than the classic positive and negative examples, a dataset is composed of a subset of positive examples $\mathcal{P}$ and a set of unlabelled examples $\mathcal{U}$, which is assumed to contain both positive and negative examples \cite{elkan2008learning}. This paradigm suitable when positive instances are a priority but labelling many instances is impractical or very expensive, such as in gene prioritisation or other bioinformatics tasks.

In bioinformatics, PU Learning has been explored in varied tasks \cite{li2022positive}. Zheng et al. \cite{zheng2019ddi} employed PU Learning with drug-drug interaction data to improve the identification of dangerous adverse reactions in patients with multiple medications. Lan et al. \cite{lan2016predicting} tackled the discovery of drug-target pairs using PU Learning. Kılıç and Mehmet \cite{kilicc2012positive} explored PU Learning to derive knowledge on protein-protein interaction networks, and Song et al. \cite{song2021inferring} used PU Learning to predict sequence-function relationships in large-scale proteomics data. Most recently, PU Learning has shown promising results in discovering toxin-degrading enzymes \cite{zhang2024discovery} or predicting the secreted proteins in human body fluids for biomarker identification of diseases \cite{he2024multi}. 

Learning from PU data is not trivial and PU Learning encompasses different strategies, such as biased learning \cite{liu2003building, sellamanickam2011pairwise}, incorporation of class prior knowledge \cite{elkan2008learning}, or the so-called \textit{two-step} methods \cite{bekker2020learning}; the latter are the focus of this work. 

Two-step methods are based around a core idea: training a binary classifier under the assumption that all unlabelled examples can be considered negative introduces label noise, hinders performance and breaks assumptions of PU tasks; this is a limitation in Magdaleno et al.'s work \cite{vega2022machine}, as discussed in Section \ref{sec:existing}. Alternatively, two-step methods propose training a binary classifier using the positive examples $\mathcal{P}$ and a subset $\mathcal{RN} \subset \mathcal{U}$ of ``reliable negatives'' extracted from the set of unlabelled examples. If the reliable negatives are correctly identified, even if scarce in number, a binary classifier can be trained with minimal label noise and respecting the PU tasks assumptions, increasing predictive performance. As such, PU learning is of particular interest in bioinformatics and biomedicine \cite{li2022positive}.

\begin{figure}[b]
    \centering
    \includegraphics[width=.80\textwidth]{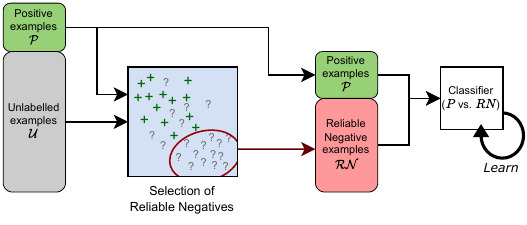}
    \caption{High-level structure of a two-step PU Learning technique.}
    \label{fig:pulearning}
\end{figure}

Figure \ref{fig:pulearning} depicts the classic template for two-step PU Learning methods, with the Unlabelled set $\mathcal{U}$ being distilled into a smaller set of Reliable Negatives $\mathcal{RN}$. Although the number of training examples is reduced in the process, the quality of training data greatly increases by minimizing label noise; ultimately, this improves the efficacy and efficiency of training and obtains classifiers with significantly higher predictive performance in PU data scenarios \cite{elkan2008learning}.

\section{The Proposed PU Learning Method}
\label{sec:proposal}

This section details the PU Learning method designed to improve the discovery of DR-related genes among ageing-related genes. Instead of treating all unlabelled examples (genes without known relation to DR) as negative examples in training, breaking the assumptions of the knowledge discovery task, we propose a training strategy where the model learns from a refined training set of positive and reliable negative examples extracted from the set of unlabelled genes. 

Our PU Learning methodology is a similarity-based method with labelling criteria inspired by a classic nearest neighbour classification (KNN) \cite{1053964}, and can be categorized as a two-step, prior-free method in Bekker and Davis' PU Learning taxonomy \cite{bekker2020learning}. As other methods in this category, we assume ageing-related data respects two assumptions: smoothness of the positive class (DR-related genes exhibit similarities in their biological features) and separability (non-DR-related genes differ from DR-related ones in their biological features). \hl{We also assume \textit{Selected Completely at Random} (SCAR) as the underlying labelling mechanism of the positive class: the set of labelled positive examples $\mathcal{G}_{\text{DR}\cap\text{AGE}}$ is an i.i.d. sample of the set of all positive examples $\mathcal{G}_{\text{DR}\cap\text{AGE}^+}$; this is a popular and reasonable assumption in bioinformatics tasks \mbox{\cite{teisseyre2024verifying}}}.

Given a dataset $\mathcal{D}$ composed of a set $\mathcal{P}$ of genes with known DR relatedness (positive examples) and a set $\mathcal{U}$ of genes without known DR relatedness (unlabelled examples), where each gene is represented with a set of biological features $\mathcal{F}$, a set $\mathcal{RN}$ of genes unlikely to be DR-related (reliable negatives) is obtained as follows:

\begin{enumerate}

    \item For each pair of genes $(g_i, g_j)$, a similarity metric is computed using their biological feature vectors $x_g = (f_1, ..., f_{|\mathcal{F}|})$. \hl{Due to the high degree of feature sparsity of the datasets involved and the fact that the datasets have only binary features (see Section \mbox{\ref{sec:datasets}}), we choose to use the Jaccard Measure $J(x_{g_i}, x_{g_j}$) \mbox{\cite{jaccard1901etude, tanimoto1958elementary}} over other options such as the Euclidean or cosine distances.} Specifically, the Jaccard similarity between two ageing-related genes represented by vectors of binary biological features is computed as:
    \begin{equation}
        \label{equation:Jaccard}
        J(x_{g_i}, x_{g_j}) = \frac{\sum_{k=1}^{|\mathcal{F}|} x_{g_i}[k] * x_{g_j}[k]}{\sum_{k=1}^{|\mathcal{F}|} x_{g_i}[k] + \sum_{k=1}^{|\mathcal{F}|} x_{g_j}[k] - \sum_{k=1}^{|\mathcal{F}|} x_{g_i}[k] * x_{g_j}[k]}
    \end{equation}
    \noindent where the numerator counts features with value 1 in both genes' feature vectors and the denominator counts features with value 1 in only one of the two genes' feature vectors.
    \item The set of reliable negatives $\mathcal{RN}$ is initialized with no elements. 
    \item For each unlabelled gene $g_i$ in the training set:
    \begin{enumerate}
        \item Find the $k$ training gene closest to $g_i$ (its $k$ nearest neighbours) based on their pairwise Jaccard similarities.
        \item Two conditions are checked:
            \begin{itemize}
                \item Whether the single closest gene to $g_i$ is unlabelled (no known DR relation).
                \item Whether the proportion of genes without known relation to DR among the top $k$ nearest neighbours of $g_i$ is higher than a set threshold $t$.
            \end{itemize}

        \item If these two conditions are met, the gene $g_i$ is added to the set of reliable negatives.
    \end{enumerate}
\end{enumerate}

\begin{figure}[]
    \centering
    \includegraphics[width=\textwidth]{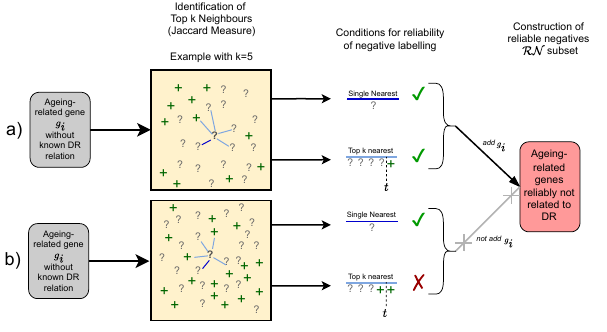}
    \caption{Similarity-based reliable negative selection of the proposed PU Learning algorithm. \hl{The threshold $t$ is the minimum proportion of unlabelled examples among the $k$ nearest neighbours of an unlabelled example required to consider it a reliable negative ($k$ and $t$ are tunable hyperparameters; in this example, $k=5$ and $t=0.8$)}. Two different cases are shown: in case (a) the two conditions for a reliable negative are met, i.e. the gene's nearest neighbour and  >80\% of its $k$ nearest neighbours are unlabelled; the gene is confidently not related to DR and is added to the set of reliable negatives. In case (b), the latter condition is not met; since the gene is not dissimilar enough to known DR-related genes, the gene is not added as a reliable negative, avoiding potential label noise during the training of the classifier in the second step of the PU Learning.}
    \label{fig:knnpu}
\end{figure}

Algorithm \ref{alg:knn} contains the logic and Figure \ref{fig:knnpu} a visual depiction of the process above, constituting the first step of the two-step PU Learning method. The result is a set of reliable negatives $\mathcal{RN}$ with very high confidence of not being DR-related. Formally, $\mathcal{RN}$ should verify:

\begin{equation}
\begin{gathered}
    \mathcal{RN} \subset \mathcal{G}_{\text{AGE}} \setminus \mathcal{G}_{\text{DR}\cap\text{AGE}^+} \\
    \mathcal{RN} \cap \mathcal{G}_{\text{DR}\cap\text{AGE}^+} = \emptyset
\end{gathered}
\end{equation}

\noindent therefore respecting the assumptions of the discovery task, as laid out in Section \ref{fig:task} and unlike the naive approach of treating all unlabelled examples as negatives.

\begin{algorithm}[h]
\footnotesize
\caption{Reliable Negatives selection by Nearest Neighbours}
\begin{algorithmic}[5]

\Require
\Statex $D$: Set of training examples 
\Statex $k$: Number of nearest neighbours
\Statex $t \in [0.5,1]$: Threshold
\Statex $F$: Features to use
\Ensure
\Statex \hl{$RN$: Set of reliable negative training examples}
\item[]
\Function{\textbf{ReliableNegatives}}{$P, U, k, t, F$}
\State $P, U \gets P, U$ considering only features F
\State $D \gets P \cup U$
\State Initialize similarity matrix $S \in \mathbb{R}^{|D| \times |D|}$
\ForAll{(\textbf{$x_i$}, \textbf{$x_j$}); \textbf{$x_i$}, \textbf{$x_j$} $\in D$} \Comment{Compute Jacard distance for all pairs of examples} 
\State $S_{ij} \gets$ J(\textbf{$x_i$},\textbf{$x_j$}) \Comment{\hl{Can cache} S to avoid re-calculations of S$_{ij}$ across ReliableNegatives calls}
\EndFor
\State $RN \gets \emptyset$
\ForAll{$x_i \in U$}
\State $T_k \gets$ Top $k$ examples $x_j \in D \setminus x_i$ with highest similarity $S_{ij}$ to $x_i$
\State $x_{max\_sim} \gets$ The example $x_j \neq x_i$ with highest similarity $S_{ij}$ to $x_i$
\LComment{If the \% of unlabelled examples in the $Top_k$ nearest neighbours of $x_i$ exceeds a certain proportion and the single nearest neighbour of $x_i$ is unlabelled, add $x_i$ to the reliable negatives set}
\If {$\dfrac{|T_k \cap U|}{|T_k|} \geq t \And x_{max\_sim} \in U$} 
\State $RN \gets RN \cup {x_i}$ 
\EndIf 
\EndFor

\Return $RN$
\EndFunction
\end{algorithmic}
\label{alg:knn}
\end{algorithm}

At the second and final step, $\mathcal{P}$ and $\mathcal{RN}$ are used to train the final model $\Phi$ with minimal label noise, improving data quality and the effectiveness and efficiency of the model learning.

This PU Learning method can be easily integrated with any classifier to predict the DR-relatedness of genes without known DR-relation, as this method is classifier-agnostic. Algorithm \ref{alg:cv+pul} shows the integration of our PU Learning method inside a standard nested cross-validation pipeline. \hl{Following the state-of-the-art for this task \mbox{\cite{vega2022machine}}, we perform a standard 10x5 nested-cross validation, where in each iteration of the outer loop, an inner loop is applied to the training set to search over the set of PU Learning hyperparameter combinations ($k$, $t$) detailed in \mbox{\ref{apx:hyperparameters}}.}

\begin{algorithm}[]
\footnotesize
\begin{algorithmic}[5]
\Require
\Statex \textbf{$D$:} Set of examples
\Statex \textbf{$k$, $t$:} PU Learning hyperparameters
\Statex \textbf{$n_f$}: No. features to use in KNN
\item[]
\State Split $D$ into 10 folds ${D_1, ..., D_{10}}$ \Comment{Outer CV}
\For{$i=1, \dots, 10$}
    \State $D_{Test} \gets D_{i}$
    \State $D_{Train} \gets D \setminus D_{i}$

    \State $AUC_{best} \gets 0$
    \ForAll{hyperparameters combination \hl{$k,t$}}
        \State Split $D_{Train}$ into 5 folds ${D_{Train_1}, ..., D_{Train_5}}$ 
        \For{$i=1, \dots, 5$}
            
            \State $D_{Val} \gets D_{Train_i}$
            \State $D_{Learn} \gets D_{Train} \setminus D_{Train_i}$
            \State Train Model with $D_{Learn}$
            \State $F \gets$ Top $n_f$ features in with highest Gini Importance in Model
            \State $P, U \gets$ Positive and Unlabelled examples of $D_{Learn}$
            
            \State $RN \gets$ \textbf{ReliableNegatives}($P$, $U$, $k$, $t$, $F$)
            \State Train Model with $P$ and $RN$ \Comment{Using all original features}
            \State Predict $D_{Val}$ with Model
        \EndFor
        \If{F1$_{Model}$ $>$ F1$_{Best}$} \Comment{Based on average AUC of 5 validation sets}
        \State $k_{best} \gets k$, $t_{best} \gets t$,  $F1_{best} \gets F1_{Model}$
        \EndIf
    \EndFor
    \State Train Model with $D_{Train}$ and hyperparams $H_{best}$
    \State $F \gets$ Top $n_f$ features in with highest Gini Importance in Model
    \State $P, U \gets$ Positive and Unlabelled examples of $D_{Train}$
    \State $RN \gets$ \textbf{ReliableNegatives}($P$, $U$, $k_{best}$, $t_{best}$, $F$)
    \State Train Model with $P$ and $RN$ \Comment{Using all original features}
    \State Predict $D_{Test}$ using Model
    \State \Comment{Final performance is average F1, AUC and G.Mean of Model in the 10 test sets}
    \EndFor
\end{algorithmic}
\caption{Integration of PU Learning in Cross Validation}
\label{alg:cv+pul}
\end{algorithm}

To increase the robustness of the method, we include an optional step in the training pipeline (see lines 14-16, 22-24 of Algorithm \ref{alg:cv+pul}). Many of the feature sets 
 of biological entities or genes are highly dimensional, and a similarity-based technique could make the PU Learning vulnerable to the \textit{curse of dimensionality}, i.e. the similarity metric would not be informative due to the low signal-to-noise ratio arising from the high number of uninformative features \cite{altman2018curse}. To solve this for high-dimensional feature sets we add the option of, initially, training a classifier of choice, and then selecting the feature set $\mathcal{F}$ to be used in the nearest neighbours algorithm as the $n_f$ most important features for that classifier, according to a feature importance measure. Using a model-based filter feature selection for a subsequent algorithm is a well-studied and efficient method compared to more complex feature selection methodologies \cite{grabczewski2005feature, ratanamahatana2003feature}.
\section{Experimental Setup}
\label{sec:setup}

This section covers the experimental setup used to evaluate the PU Learning methodology, compare it to existing approaches, and generate the ranking of promising candidate genes for DR-relatedness. We discuss the feature sets and classifiers used, details of the training and evaluation pipelines, and other implementation peculiarities. 

\subsection{Features and Classifiers}

\label{sec:datasets}

In Section \ref{sec:task} we outlined that each ageing-related gene is represented as a vector of biological features, as well as a binary class label that indicating whether it has a known relation with DR (i.e. whether it belongs to $\mathcal{G}_{\text{DR}\cap\text{AGE}}$). 

In this work, we focalize on the two feature sets originally found to be most relevant for the prediction of DR-relatedness by Magdaleno et al. \cite{vega2022machine}: PathDIP and GO features. Table \ref{tab:datasets} shows statistics of the two constructed datasets (one per feature type), namely the number of features, the number of genes (examples) in each class and the feature sparsity (percentage of 0-valued features) for each feature type. These two learning scenarios are created as follows: 

\begin{itemize}
    \item Using PathDIP \cite{rahmati2017pathdip}, each gene is represented by binary features indicating whether or not it belongs to an specific PathDIP pathway; PathDIP, on its own, integrates information from multiple database sources such as Bio-Carta, REACTOME, UniProt, etc. \hl{For instance, if a gene has value $1$ for the feature \textit{KEGG.2}, then that gene has experimental evidence of relation to the \textit{KEGG.2} pathway, which regulates \textit{animal autophagy}.}
    \item Using GO \cite{ashburner2000gene}, each gene is represented by binary features indicating relation to a specific GO term or any of its descendants. \hl{For example, if a gene has value $1$ for the feature \textit{GO:0055114}, then that gene has experimental evidence of being related to the \textit{Oxidation-reduction process}.}
\end{itemize}

\begin{table}[h]
\centering
\caption{Basic statistics of the two used datasets (feature types), with PathDIP \cite{rahmati2017pathdip} and GO \cite{ashburner2000gene} features. \hl{Among all ageing-related genes (i.e. $|\mathcal{G}_{\text{AGE}}|$ is the number of instances), the known DR-related genes ($\mathcal{G}_{\text{DR}\cap\text{AGE}} \subset \mathcal{G}_{\text{AGE}}$) constitute the subset of known positive instances or examples.} 
}
\label{tab:datasets}
\resizebox{.7\textwidth}{!}{%
\begin{tabular}{lrrrr}
\hline
\multicolumn{1}{c}{\textbf{Feature Set}} & \multicolumn{1}{c}{\textbf{ Features}} & \multicolumn{1}{c}{\textbf{\begin{tabular}[c]{@{}c@{}} \hl{Ageing-related}\\ \hl{Genes ($|\mathcal{G}_{\text{AGE}}|$)}\end{tabular}}} & \multicolumn{1}{c}{\textbf{\begin{tabular}[c]{@{}c@{}} \hl{Known DR-Related} \\ \hl{Genes ($|\mathcal{G}_{\text{DR}\cap\text{AGE}}|$)} \end{tabular}}} & \multicolumn{1}{l}{\textbf{\begin{tabular}[c]{@{}c@{}}Feature \\ Sparsity (\%)\end{tabular}}} \\ \hline
PathDIP                                  & 1,640                                    & 986                                                                                           & 110                                                                                               & 98.39\%                                                                                       \\
GO                                       & 8,640                                    & 1,124                                                                                         & 114                                                                                               & 98.46\%                                                                                       \\ \hline
\end{tabular}%
}
\end{table}

We followed Magdaleno et al.'s work to retrieve ageing-related genes, their features and their known DR-relatedness. To identify ageing-related human genes ($\mathcal{G}_{\text{AGE}}$), the GenAge database \cite{de2004genage} was queried for genes that affected ageing phenotype or longevity if modulated in model organisms, and the obtained genes were mapped to their human orthologs using the OMA Orthology database \cite{altenhoff2021oma} to obtain the final set of human ageing-related genes. The GenDR database \cite{tacutu2018human} was queried to obtain DR-related genes in model organisms as those that affected the effectiveness of the DR-mediated ageing process in at least one wet lab experiment. Again, these DR-related genes were mapped to their human orthologs using the OMA Orthology database. The ageing-related genes were labelled as DR-related if they belonged to this set of DR-related genes, and PathDIP and GO features were retrieved for all ageing-related genes. \hl{For example, the ageing-related gene \textit{MDH1} is labelled as positive (DR-related) because there is firm evidence it activates downstream targets of DR like SIR2 \mbox{\cite{easlon2008malate}}, while the ageing-related gene \textit{PRKAB1} is unlabelled because there is an absence of evidence linking it to DR.}

\hl{With regard to the classifiers used, our PU Learning method is classifier-agnostic, but we maintain the use of decision tree-based ensemble methods as done by Magdaleno et al. \mbox{\cite{vega2022machine}}; this is both to ensure fairness in the comparison with \mbox{\cite{vega2022machine}}, and because tree-based ensembles are the State of the Art for tabular data, over options like Logistic Regression, Support Vector Machines or Neural Networks \mbox{\cite{grinsztajn2022tree, shwartz2022tabular}}. In particular, we employed CAT and BRF, the two classifiers with highest performance in the state-of-the-art approach for the problem \mbox{\cite{vega2022machine}}:}
\begin{itemize}
    \item CatBoost (CAT) \cite{prokhorenkova2018catboost} is a boosting-based ensemble classifier: each base learner is trained sequentially with instance weights determined by the errors of previous base learners in the sequence, progressively reducing the bias in the predictions \cite{buhlmann2012bagging}.
    \item Balanced Random Forest (BRF) \cite{chen04using} is a bagging-based ensemble classifier: each base learner is trained independently using a bootstrap sample of the original data, and the high predictive accuracy is obtained through a reduction in variance of the errors, improving the unstable and inaccurate estimations of the weak base learners in isolation \cite{buhlmann2012bagging}.
\end{itemize} 

\subsection{Evaluation Details}

In order to evaluate our proposed PU Learning method for the discovery of new candidate DR-related genes, we performed a dual evaluation, both on the surrogate binary classification task (predictive performance and computational cost) and on the proposed candidate ranking of the most promising novel DR-related genes.

\hl{To evaluate predictive performance we measure, for the existing (non-PU) approach \mbox{\cite{vega2022machine}} and our proposed PU Learning method, three relevant ML performance metrics: the F1 Measure of the positive class, the Geometric Mean (G. Mean), and the AUC-ROC \mbox{\cite{japkowicz2011evaluating}}, such that:} 
\begin{equation}
\begin{aligned}
\colorbox{ChangesColor}{$F1 = \frac{2*Precision*Recall}{Precision+Recall} \;\;\;$} & \colorbox{ChangesColor}{$\;\;\; G.Mean = \sqrt{Sensitivity*Specificity}$} \\[3ex]  
\omit\rlap{\colorbox{ChangesColor}{AUC-ROC $=$ Area Under the $Sensitivity\;vs.\;(1-Specificity)$ Curve}} 
\end{aligned}
\end{equation}

\noindent \hl{where Precision is the proportion of instances annotated as positives (DR-related) in the data among all instances predicted as positives,  
Recall (or Sensitivity) is the proportion of instances predicted as positives among the set of all instances annotated as positives in the data,
and Specificity is the proportion of instances predicted as negatives (non-DR-related) among the set of all instances annotated as negatives in the data.}

\hl{G. Mean and AUC-ROC are broadly used in PU Learning works for their global performance overview across both classes \mbox{\cite{saunders2022evaluating}}, and are used in the only existing work on DR-related genes identification \mbox{\cite{vega2022machine}}. However, AUC-ROC can be unreliable in PU tasks on datasets with class imbalance \mbox{\cite{bekkar2013evaluation}}, but we keep it for fairness of comparison with \mbox{\cite{vega2022machine}}. We avoid the use of accuracy due to its inadequacy in scenarios with class imbalance \mbox{\cite{bekkar2013evaluation}}, and we favour the use of F1 Score over its components Recall and Precision for this same reason.}

\hl{We use the F1 Measure as the main performance indicator, as it is the most popular metric for PU Learning in the literature \mbox{\cite{saunders2022evaluating}}}. It is worth noticing that, because we utilized genuine PU data (i.e. we do not know the real labels of any of the unlabelled examples), all three metrics are an estimation of their real values: they evaluate the performance of the classifier predicting whether the example is a known positive ($Pr(g_i \in \mathcal{G}_{\text{DR}\cap\text{AGE}})$) rather than whether it is a real positive ($Pr(g_i \in \mathcal{G}_{\text{DR}\cap\text{AGE}^+})$). In this regard, the estimation of the F1 Measure of the positive class has desirable properties: Elkan and Noto \cite{elkan2008learning} showed that (1) it will be a strict underestimation of the real F1 Measure, and (2) it will differ from the real F1 Measure by a constant factor, making it suitable for confidently comparing classifiers on discovery tasks. 

\hl{With respect to the analysis of computational cost, we track the greenhouse gas emissions (grams of carbon dioxide equivalent or gCO$_2$e) of the DR-related gene identification pipelines of the existing (non-PU) method \mbox{\cite{vega2022machine}} and our PU-based Learning method, across the nested cross-validation training and inference pipeline, using \textit{codecarbon} \mbox{\cite{benoit_courty_2024_11171501}}.}

We also performe two qualitative evaluations of our best learned model: first, we analyzed the most important features to predict DR-relatedness according to the classifier, based on the Mean Decrease in Impurity \cite{louppe2013understanding}, and compared those features with the most important features identified in \cite{vega2022machine}. \hl{Second, we produce a ranking of the most promising ageing-related genes for novel DR-relatedness as predicted by our model. As formalized in Eq. \mbox{\ref{eq:topcandidates}}, these most promising genes (i.e. those most likely to belong to $\mathcal{G}_{\text{DR}\cap\text{AGE}^+} \setminus \mathcal{G}_{\text{DR}\cap\text{AGE}}$) are defined by the model as those unlabelled examples (genes without known DR-relatedness) that are predicted as belonging to the positive class with the highest probability (see Eq. \mbox{\ref{eq:aprox}}). We use online resources (\textit{Pubmed}, \textit{Google Scholar}) to search for research linking a gene or its encoded protein to biological mechanisms potentially related to DR, 
for the most promising genes reported by our PU-based method and the existing approach by Magdaleno et al. \mbox{\cite{vega2022machine}}}.

\subsection{Implementation Details}

\hl{This section covers technical details of the experiments performed to measure the performance of our PU Learning method and the existing (non-PU) Learning method from \mbox{\cite{vega2022machine}}}: 

\begin{itemize}
    \item \hl{We implemented and performed all our experiments, using both the existing (non-PU) approach and our PU Learning approach, in a common Python framework}. CatBoost and Balanced Random Forest are integrated through the \textit{catboost}\footnote{\href{https://catboost.ai/en/docs/}{https://catboost.ai/en/docs/}} and \textit{imbalanced-learn}\footnote{\href{https://imbalanced-learn.org/stable/}{https://imbalanced-learn.org/stable/}} packages, respectively. This framework\footnote{\href{https://github.com/Kominaru/DR\_Gene\_Prediction\_XofN\_PUL}{https://github.com/Kominaru/DR\_Gene\_Prediction\_XofN\_PUL}} is publicly made available to the scientific community for transparency and further experimentation with these methods. 
    \item For result reliability, the evaluation (nested cross-validation and ranking of candidate genes) of each method is averaged over 10 executions with random state seeds ($14$, $33$, $39$, $42$, $727$, $1312$, $1337$, $56709$, $177013$, $241543903$).
    \item For the existing (non-PU) method, we replicated the original hyperparameter space and grid search procedure performed in \cite{vega2022machine}. For our PU Learning method, we tuned the PU Learning hyperparameters as described in Section \ref{sec:proposal}. \hl{\mbox{\ref{apx:hyperparameters}} details the values of tunable and non-tunable hyperparameters of all methods used in our experiments.}
    \item We run all our experiments in a dedicated machine with 32GB RAM, i9-10980 Intel\copyright CPU, NVIDIA RTX 3070 GPU, and Windows 10 OS.

\end{itemize}
\section{Results}
\label{sec:results}

\hl{This Section covers the results of our PU Learning approach for detecting novel DR-related genes among ageing-related genes comparing its resuts with the results of the only existing approach for this task, a non-PU methodology recently introduced in \mbox{\cite{vega2022machine}}}. We compare the two approaches' predictive performance results, computational cost, most important predictive features, and most promising candidate genes for novel DR-relatedness.

\subsection{Results of Computational Experiments}
\label{sec:mlperformance}
Table \ref{tab:mlresults} shows the predictive performance results (for the three metrics introduced in Section \ref{sec:setup}) of our PU Learning method and the existing, non-PU method \cite{vega2022machine}. For both methods, we consider two classifiers (CatBoost and Balanced Random Forest) and two biological feature sets (PathDIP pathways and GO terms). 

The best results for each metric (AUC-ROC, G. Mean and F1 Score) are all obtained with our PU Learning method, showing statistical difference on two-tailed t-tests with $\alpha=0.05$ against all other results. We highlight the relevance of the F1 Score, maximized by the PU Learning approach using \{PathDIP, CAT\}, as it is the most informative performance metric in PU Learning tasks \cite{saunders2022evaluating}. As such, overall, our PU Learning method exhibits a stronger predictive performance than the original non-PU Learning approach by Magdaleno et al. \cite{vega2022machine}.

\begin{table}[h]
\caption{Predictive performance of the original non-PU Learning method and our proposed PU Learning method in the identification of DR-related genes among ageing-related genes, considering two feature sets and two classifiers. Results represent average performance across 10 executions of the nested cross-validation procedure. For each metric, the best result is bolded and a dagger (†) represents statistical significance against all other results (in the other 7 rows) on two-tailed t-tests with $\alpha = 0.05$.}
\label{tab:mlresults}
\renewcommand{\arraystretch}{1.5}
\centering
\resizebox{.7\textwidth}{!}{%
\begin{tabular}{lllrrr}
\hline
\multirow{2}{*}{\textbf{Method}} &
  \multirow{2}{*}{\textbf{Features}} &
  \multirow{2}{*}{\textbf{Classifier}} &
  \multicolumn{3}{c}{\textbf{Performance (avg of 10 runs)}} \\ \cline{4-6} 
 &
   &
   &
  \multicolumn{1}{l}{\textbf{AUC-ROC}} &
  \multicolumn{1}{l}{\textbf{G. Mean}} &
  \multicolumn{1}{l}{\textbf{F1 Score}} \\ \hline
\multirow{4}{*}{Original}    & \multirow{2}{*}{PathDIP} & CAT & 0.829           & 0.717           & 0.522           \\
                             &                          & BRF & 0.825           & 0.752           & 0.450           \\
                             & \multirow{2}{*}{GO}      & CAT & 0.832           & 0.654           & 0.463           \\
                             &                          & BRF & 0.827           & 0.755           & 0.377           \\ \cline{1-3}
\multirow{4}{*}{PU Learning} & \multirow{2}{*}{PathDIP} & CAT & 0.829           & 0.750           & \textbf{0.537†} \\
                             &                          & BRF & 0.815           & 0.728           & 0.381           \\
                             & \multirow{2}{*}{GO}      & CAT & \textbf{0.838†} & 0.726           & 0.491           \\
                             &                          & BRF & 0.829           & \textbf{0.763†} & 0.380           \\ \hline
\end{tabular}%
}

\end{table}

Comparing the non-PU Learning and PU Learning approaches in each isolated \{Features, Classifier\} scenario, in both GO scenarios (\{GO, CAT\} and \{GO, BRF\}) the PU Learning approach outperformed the original non-PU method for all three performance metrics. In the \{PathDIP, CAT\} scenario, our PU-based model outperformed the non-PU Learning one in terms of F1 Score and G.Mean, exhibiting equivalent AUC-ROC. Only in one scenario (\{PathDIP, BRF\}) the usage of the PU Learning approach did not benefit the predictive performance; in this scenario, it seems that the benefits of a more reliable negative example labelling for training may not have compensated for the performance penalty that reducing the number of available training examples causes to a bagging-based method like BRF.

Overall, the best performance among all options in Table \ref{tab:mlresults} was observed for the PU Learning approach in the \{PathDIP, CatBoost\} scenario, since it shows the statistically best result for F1, the most important metric -further analyzed in \ref{apx:f1}-, and exhibits competitive results in the other two metrics (AUC-ROC and G. Mean). As such, owing to an overall higher performance in these experiments involving two classifiers, two feature sets and three performance metrics, we can conclude that the proposed PU Learning approach is superior to state-of-the-art non-PU Learning methods in the the identification of DR-related genes among ageing-related genes, and therefore should produce more reliable rankings of top candidates for DR-relatedness among genes without known relationship to DR. 

\begin{figure}[!ht]
    \centering
    \includegraphics[width=0.8\linewidth]{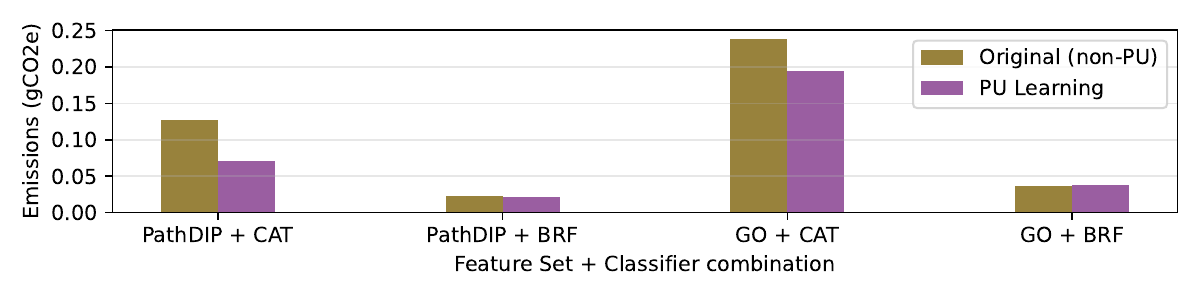}
    \caption{\hl{Comparison of the computational cost (measured in grams of carbon dioxide equivalent (gCO$_2$e), lower is better) of Magdaleno et al.'s original non-PU approach and our proposed PU Learning-based approach for identification of new DR-related genes. Results are averaged over 10 complete executions of the nested cross-validation, involving training and inference procedures}.}
    \label{fig:emissions}
\end{figure}

\hl{With respect to the computational cost, Figure \mbox{\ref{fig:emissions}} shows the average CO$_2$ emissions of the combined training and inference phases of the existing non-PU approach by Magdaleno et. al \mbox{\cite{vega2022machine}} and our proposed PU-based methodology. It can be observed that, despite the additional steps required for the PU Learning-based labelling of examples, this does not worsen the computational overhead of the identification pipeline. In fact, in the \{PathDIP, CAT\} scenario, where both the non-PU and our PU-based approach obtain highest predictive performance, our PU Learning approach can identify novel DR-related genes requiring $\sim40\%$ less greenhouse gas emissions. This is because, even if additional computations are required to select the reliable negative examples for the classifier, the resulting refined training set has fewer but better quality negative examples in it, prompting a more cost-effective learning of the model.}

\subsection{Analysis of Most Important Predictive features}

Table \ref{tab:featureimportances} shows the 5 most important features in the best models learnt for predicting DR-relatedness, i.e. the top-5 reported in \cite{vega2022machine} for the best model with the existing non-PU-Learning approach (left), and the top-5 in the best model learned with our PU Learning approach (right). This is a well-controlled comparison, as both sets were obtained in the \{PathDIP, CatBoost\} scenario. These feature importance values, as done in \mbox{\cite{vega2022machine}}, were computed using the Mean Decrease in Impurity, which measures to what extent the splits with a given feature decrease the Gini Index across the trees in the ensemble \cite{louppe2013understanding}. 

\begin{table}[ht]
\caption{\hl{The 5 most important features used to predict DR-relatedness using Magdaleno et al.'s non-PU Learning method \mbox{\cite{vega2022machine}} and our PU Learning algorithm, using PathDIP as feature set and CatBoost as classifier. The feature importance used is the Mean Decrease in Impurity based on the Gini Index. Each feature importance score was averaged across 10 executions of the nested 10x5 Cross Validation (CV) procedure, and it was normalised to the [0, 100] range. In each 10x5 CV, the feature importance is computed and averaged for each outer fold after the inner Cross Validation optimizes $k$ and $t$. Features common to both approaches are shown in italics.}}
\label{tab:featureimportances}
\renewcommand{\arraystretch}{1.5}
\centering
\resizebox{\textwidth}{!}{%
\begin{tabular}{llrlllr}
\hline
\multicolumn{3}{c}{\textbf{Original non-PU-Learning method}}                                                                                                                                                                         &  & \multicolumn{3}{c}{\textbf{KNN-based PU-Learning method}}                                                                                                                                                                  \\ \cline{1-3} \cline{5-7} 
\multicolumn{1}{c}{\textbf{Feature}}                                         & \multicolumn{1}{c}{\textbf{Definition}}                                                   & \multicolumn{1}{c}{\textbf{Score}} &  & \multicolumn{1}{c}{\textbf{Feature}}                                        & \multicolumn{1}{c}{\textbf{Definition}}                                                    & \multicolumn{1}{c}{\textbf{Score}} \\ \hline
\textit{\begin{tabular}[c]{@{}l@{}}KEGG.2 \\ (map04140)\end{tabular}}        & Autophagy - animal                                                                        & 100.00                             &  & \textit{\begin{tabular}[c]{@{}l@{}}WikiPathways.37\\ (WP2884)\end{tabular}} & NRF2                                                                                       & 100.00                             \\
\textit{\begin{tabular}[c]{@{}l@{}}KEGG.30 \\ (map04213)\end{tabular}}       & \begin{tabular}[c]{@{}l@{}}Longevity regulating\\ pathway - multiple species\end{tabular} & 45.35                              &  & \textit{\begin{tabular}[c]{@{}l@{}}KEGG.2 \\ (map04140)\end{tabular}}       & Autophagy - animal                                                                         & 72.52                              \\
\begin{tabular}[c]{@{}l@{}}NetPath.23 \\ (Pathway\_BDNF)\end{tabular}        & \begin{tabular}[c]{@{}l@{}}Brain-derived \\ neurotrophic factor\end{tabular}              & 38.95                              &  & \textit{\begin{tabular}[c]{@{}l@{}}KEGG.30 \\ (map04213)\end{tabular}}      & \begin{tabular}[c]{@{}l@{}}Longevity regulating \\ pathway - multiple species\end{tabular} & 56.32                              \\
\begin{tabular}[c]{@{}l@{}}REACTOME.10 \\ (R-HSA-8953897)\end{tabular}       & \begin{tabular}[c]{@{}l@{}}Cellular responses \\ to external stimuli\end{tabular}         & 37.79                              &  & \begin{tabular}[c]{@{}l@{}}WikiPathways.57 \\ (WP534)\end{tabular}          & \begin{tabular}[c]{@{}l@{}}Glycolysis and \\ gluconeogenesis\end{tabular}                  & 47.85                              \\
\textit{\begin{tabular}[c]{@{}l@{}}WikiPathways.37 \\ (WP2884)\end{tabular}} & NRF2                                                                                      & 34.88                              &  & EHMN.11                                                                     & \begin{tabular}[c]{@{}l@{}}Fructose and \\ mannose metabolism\end{tabular}                 & 43.97                              \\ \hline
\end{tabular}%
}
\end{table}

In Table \ref{tab:featureimportances}, three features ranked in the top-5 features for both approaches: KEGG pathway map04140 (Autophagy -- animal), KEGG pathway map04213 (Longevity regulating pathway -- multiple species), and WikiPathway WP2884 (NRF2). While ``Longevity regulating pathway -- multiple species'' is a very broad KEGG pathway without clear relation to DR, there is good support in the literature for pathways map04140 and WP2884: autophagy inhibition tends to attenuate the anti-ageing effects of Calorie Restriction (CR), and a review on CR and autophagy has concluded that there is strong evidence that fasting and CR promote autophagy in a wide variety of tissues and organs \cite{Bagherniya2018fasting}.
Nuclear factor erythroid 2-related factor 2 (NRF2) is a transcription factor with activity regulated by DR that affects the expression of several enzymes with antioxidant and detoxifying functions \cite{Vasconcelos2019Nrf2}.

Among the top-5 predictive features for the non-PU Learning approach reported in \cite{vega2022machine}, NetPath Pathway\_BDNF (Brain-derived neurotrophic factor) and Reactome R-HSA-8953897 (Cellular responses to external stimuli) are not among the top-5 features for our PU Learning method. Interestingly, it was noted in \cite{vega2022machine} that, among their reported top-5 features, only Pathway\_BDNF and R-HSA-8953897 did not show significantly different degree of occurrence between genes with and without known DR relation, so the support for these features is weaker than for the other top-5 features. This was detected by the proposed PU Learning approach, which only ranked Pathway\_BDNF and R-HSA-8953897 63rd and 34th, respectively, in terms of feature importance for predicting DR-relatedness.

Among the top-5 predictive features in our best PU Learning-based model, two are not among the top-5 features reported for the best model in \cite{vega2022machine}: WikiPathways WP534 (Glycolysis and gluconeogenesis) and EHMN.11 pathway (Fructose and mannose metabolism). Upon review of the existing scientific literature, there exists support for a significant role of these pathways in CR. Regarding WP534, in recent experiments quantifying the hepatic proteome of mice exposed to graded levels of CR (from 0\% to 40\%) for 3 months, one of the metabolic pathways most significantly stimulated by an increase in the level of CR was the glycolysis/gluconeogenesis pathway \cite{Wang2023impact}. In addition, glycolysis and gluconeogenesis were considerably up-regulated in the kidney tissue of rats undergoing CR for 6 months \cite{Chen2008identifying}. Regarding EHMN.11, in a study to investigate the successful maintenance of weight loss in people after 8 weeks of low-calorie diet where people were classified as weight maintainers or weight regainers, an analysis of subcutaneous adipose tissue gene expression showed that the the low-calorie diet caused a decrease in the fructose and mannose metabolism pathway in the weight regainer group \cite{Mutch2011distinct}.

When comparing the two sets of top-5 features identified by the two approaches in Table  \ref{tab:featureimportances}, it is worth recalling that the PU Learning-based model obtained better predictive performance than the non-PU model, as reported in Section \ref{sec:mlperformance}. Therefore, it is reasonable to consider that the top-5 features identified by the PU Learning-based model are stronger, more reliable predictors of DR-relatedness than the top-5 features reported in \cite{vega2022machine}.

\subsection{Analysis of the Most Promising Candidate DR-related Genes}

We employed the overall best method in Section \ref{sec:mlperformance} (the proposed PU learning method using PathDIP as feature set and CatBoost as classifier), to obtain the ranking of the most promising candidate ageing-related genes for novel DR-relatedness. For consistency with the evaluation scheme adopted in \cite{vega2022machine}, which uses non-PU Learning methods, we report here the top-7 genes in the obtained ranking for each method. 

\begin{table}[H!]
\caption{\hl{Top 7 candidate genes for novel DR-relatedness, i.e. genes without known DR association but with the highest likelihood of being DR related, for both the state-of-the-art non-PU Learning method and our proposed PU Learning algorithm. The DR-Probability is the output of the model for each gene, averaged across 10 executions of the 10x5 Cross Validation procedure. In each 10x5-CV, the DR-Probability is computed for the outer fold where the gene is in the test partition, after the inner Cross Validation optimizes $k$ and $t$. Genes common to both rankings are shown in italics.}}
\label{tab:candidates}
\renewcommand{\arraystretch}{1.5}
\centering
\resizebox{.7\textwidth}{!}{%
\begin{tabular}{lrllr}
\hline
\multicolumn{2}{l}{\textbf{Original non-PU-Learning method}}     &  & \multicolumn{2}{l}{\textbf{KNN-based PU-Learning method}}  \\ \cline{1-2} \cline{4-5} 
\textbf{Gene}   & \textbf{DR-Probability} &  & \textbf{Gene}   & \textbf{DR-Probability} \\ \hline
GOT2   & 0.86                    &  & \textit{TSC1}   & 0.97                    \\
GOT1   & 0.85                    &  & \textit{GCLM}   & 0.94                    \\
\textit{TSC1}             & 0.85                    &  & IRS1            & 0.93                    \\
CTH            & 0.85                    &  & PRKAB1 & 0.92                    \\
\textit{GCLM}           & 0.82                    &  & PRKAB2          & 0.90                    \\
\textit{IRS2}            & 0.80                    &  & PRKAG1          & 0.90                    \\
SENS2           & 0.80                    &  & \textit{IRS2}            & 0.90                    \\
\hline
\end{tabular}%
}
\end{table}

Table \ref{tab:candidates} shows these top-7 most promising genes as identified in \cite{vega2022machine} (left) and by our PU-learning method (right); there is an overlap of 3 genes (\textit{TSC1}, \textit{GCLM}, \textit{IRS2}) and 4 differing genes between the two approaches. Among the top-7 genes identified in \cite{vega2022machine}, the 4 genes not occurring in the list of top-7 genes identified in this work are \textit{GOT2}, \textit{GOT1}, \textit{CTH} and \textit{SESN2}, ranked 10th, 11th, 23rd and 115th by our PU learning approach, respectively. Conversely, among the top-7 genes identified by the PU learning approach, 4 candidate DR-related genes do not occur in the list of top-7 genes identified in \cite{vega2022machine}: \textit{IRS1}, \textit{PRKAB1}, \textit{PRKAB2} and \textit{PRKAG1}. In the following paragraphs we discuss how evidence from the relevant literature supports the possible DR-relatedness of these 4 candidate DR-related genes identified by our PU Learning-based method (which should identify more reliable candidate genes owing to its higher predictive performance, as discussed in Section \ref{sec:mlperformance}).

The PRKAB1 and PRKAB2 genes encode two isoforms of AMPK's regulatory $\beta$-subunit ($\beta$1 and $\beta$2) \cite{Katwan2019AMPactivated}; and PRKAG1 encodes an isoform of AMPK's regulatory subunit $\gamma$1 \cite{An2020importance}. In previous research, fasting for 24 hours increased the gene expression of AMPK $\beta$1- and $\beta$2-subunits in the hypothalamus of chicks \cite{Lei2012fastingChicks}. In addition, experiments with a mouse model of Parkinson’s disease showed that the hormone ghrelin mediated the neuroprotective effect of CR, and that the selective deletion of AMPK $\beta$1- and $\beta$2-subunits in dopamine neurons prevented ghrelin-induced AMPKD phosphorylation and neuroprotection \cite{Bayliss2016ghrelin}.

PRKAG1 has shown to have an important role in the fasting-refeeding cycle associated with DR in killifish \cite{Ripa2023refeeding}: in young killifish, the fasting-refeeding cycle triggers an oscillatory regulation pattern in the expression of genes encoding the AMPK regulatory subunits $\gamma$1 and $\gamma$2, where fasting induces $\gamma$2 and suppresses $\gamma$1, whereas refeeding induces $\gamma$1 and suppresses $\gamma$2. This regulation pattern is blunted in old age, resulting in reduced PRKAG1 expression, which leads to chronic metabolic quiescence. Transgenic killifish with sustained AMPK-$\gamma$1 avoided that metabolic quiescence, leading to a more youthful feeding and fasting response in older killifish, with improved metabolic health. Hence, Ripa et al. \cite{Ripa2023refeeding} have proposed that the selective stimulation of AMPK-$\gamma$1 could be a good strategy to reinstate the beneficial response of a late-life DR through the maintenance of a correct refeeding response.

Regarding IRS1, experiments have shown a 109\% in tyrosine-phosphorylated \textit{IRS1} in insulin-treated muscles from rats on CR by comparison with control (fed \textit{ad libitum}) rats \cite{Dean2000Calorie}, while experiments with mice lacking \textit{IRS1} showed that the \textit{IRS1}-encoded protein is not required for the CR-induced increase in insulin-stimulated glucose transport in skeletal muscle, and the absence of \textit{IRS1} did not modify any of the measured characteristic adaptations of CR \cite{Gazdag1999Calorie}.

\section{Conclusions}
\label{sec:conclusions}

\hl{
This work tackles the use of ML methods to identify new DR-related genes among ageing-related genes. The existing state-of-the-art method \mbox{\cite{vega2022machine}} treats as negative training examples (i.e. as non-DR-related genes) all genes without experimental evidence of DR-relatedness (unlabelled examples), introducing label noise to the training data and reducing the reliability of the identified candidate genes. 

To address this limitation we propose a two-step, similarity-based PU Learning methodology for gene prioritisation that creates a higher quality training set where all negative examples are reliable (confidently non-DR-related) to train the classifier. We compared our PU-based method against the state-of-the-art, non-PU-based methodology \mbox{\cite{vega2022machine}} for the task of identifying DR-related genes among ageing-related genes. 

We show that our PU Learning approach significantly ($p<0.05$) outperforms the state-of-the-art methodology \mbox{\cite{vega2022machine}} for our target task, in terms of F1 Score, G.Mean and AUC-ROC. Moreover, our method lowers the computational cost of the gene identification task by up to $\sim40\%$ in the best-performing scenario, as it generates a set of negative training examples that is smaller but has higher quality (it is more reliable), allowing a more cost-effective learning.

We use our best model (trained in the \{PathDIP, CatBoost\} scenario) to generate a ranking of candidate DR-related genes; and compare it to the ranking reported in \mbox{\cite{vega2022machine}} obtained without PU Learning. We identify several new potentially DR-related genes (e.g. IRS1, PRKAB1, PRKAB2, PRKAG1). Our new list of candidates is, from a ML standpoint, more reliable than the one curated by Magdaleno et al., since it was generated by a model with significantly higher predictive performance. Moreover, a curation of the scientific literature of these genes supports the potential relation of the top promising genes with the mechanisms of DR.

Regarding future research, we identify different avenues: (1) the validation of the DR-related role of the identified genes in wet-lab experiments, (2) the combination of multiple and other biological feature sets with our PU-based method to improve the quality of predictions, and (3) exploration of other types of classifiers to further evaluate our PU Learning method in gene prioritisation scenarios.  
}
\section*{Acknowledgements}

This research work has been funded by MICIU/AEI/10.13039/501100011033 and ESF+ (grant FPU21/05783), MICIU/AEI/10.13039/501100011033/ and
\textit{ERDF A way of making Europe} (grant PID2019-109238GB-C22), and by the Xunta de Galicia (Grant ED431C 2022/44) with the European Union ERDF funds. CITIC, as Research Center accredited by Galician University System, is funded by ``Consellería de Cultura, Educación e Universidade from Xunta de Galicia'', supported in an 80\% through ERDF Operational Programme Galicia 2021-2027, and the remaining 20\% by ``Secretaría Xeral de Universidades'' (Grant
ED431G 2023/01).

\appendix
\section{\hl{Detailed Experimental Hyperparameters}}
\label{apx:hyperparameters}
\renewcommand{\arraystretch}{1.5}
\begin{table}[h]
\caption{\hl{Full relation of classifier (non-tunable) and PU Learning (tunable) hyperparameters used in our experiments. For CAT and BRF, all hyperparameter values are identical to those in \mbox{\cite{vega2022machine}}, and hyperparameters not stated in this table use the default values in \textit{python} libraries \textit{catboost v1.2.2} and \textit{imbalanced-learn v0.12.0}.}}
\label{tab:hyperparameters}
\begin{tabular*}{.8\columnwidth}{@{\extracolsep{\fill}}llr}
\hline
\multicolumn{1}{c}{\textbf{Hyperparameter Group}} &
  \multicolumn{1}{c}{\textbf{Parameter}} &
  \multicolumn{1}{c}{\textbf{Value(s)}} \\ \hline
\multirow{2}{*}{PU Learning (tunable)} &
  No. of Neighbours &
  $k \in {3, 5, 8}$ \\
 &
  Selection Threshold &
  \begin{tabular}[c]{@{}r@{}}$k=3: t \in{\nicefrac{2}{3}, 1}$\\ $k=5: t \in{\nicefrac{4}{5}, 1}$\\ $k=8: t \in{\nicefrac{6}{8}, \nicefrac{7}{8}, 1}$\end{tabular} \\ \hline
\multirow{4}{*}{BRF} & Estimators        & $n = 500$         \\
                     & Sampling Strategy & $\omega_{us} = 1$ \\
                     & Replacement       & $r = True$        \\ \hline
CAT                  & Estimators        & $n = 500$         \\ \hline
\end{tabular*}
\end{table}
\section{\hl{Detailed F1 Score Results}}
\label{apx:f1}
\begin{figure}[h]
    \centering
    \includegraphics[width=1\linewidth]{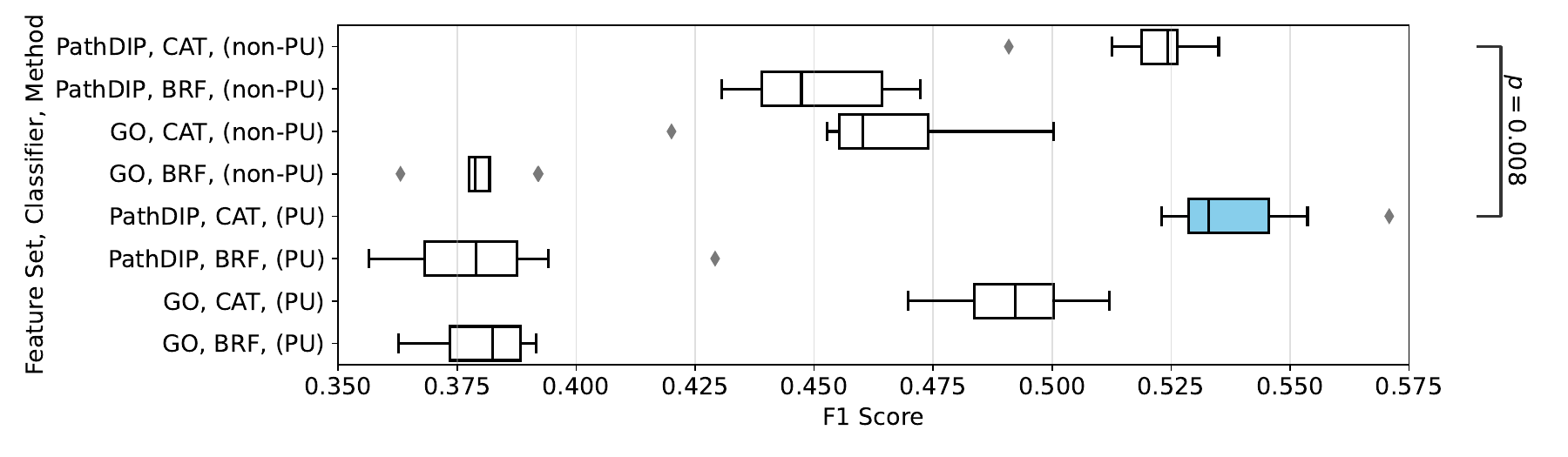}
    \caption{\hl{Details of the F1 Score across 10 executions of the nested cross-validation for the existing method \mbox{\cite{vega2022machine}} and our PU Learning-based proposal. For the best method, highlighted in blue (PU Learning on the \{PathDIP, CAT\} scenario), the $p$-value of the paired t-test against the best-performing scenario of the non-PU method is shown.}}
    \label{fig:f1detail}
\end{figure}

\printcredits

\bibliographystyle{model1-num-names-copy}

\bibliography{dr-pu}



\end{document}